\documentclass[10pt,twocolumn,letterpaper]{article}

\usepackage[pagenumbers]{cvpr} 
\usepackage{booktabs}
\usepackage{multirow}
\usepackage[sort&compress]{natbib}

\usepackage[dvipsnames]{xcolor}

\definecolor{cvprblue}{rgb}{0.21,0.49,0.74}
\usepackage[pagebackref,breaklinks,colorlinks,citecolor=cvprblue]{hyperref}

\def\paperID{*****} 
\def\confName{CVPR}
\def\confYear{2024}

\title{Evaluation of Geolocation Capabilities of Multimodal Large Language Models and Analysis of Associated Privacy Risks}

\author{Xian Zhang\\
State Key Laboratory of Information Engineering in Surveying, Mapping and Remote Sensing\\
129 Luoyu Road, Hongshan District, Wuhan University School of Information, Wuhan, China\\
{\tt\small zhangxian@whu.edu.cn}
\and
Xiang Cheng\\
State Key Laboratory of Information Engineering in Surveying, Mapping and Remote Sensing\\
129 Luoyu Road, Hongshan District, Wuhan University School of Information, Wuhan, China\\
{\tt\small 2020302142087@whu.edu.cn}
}

\begin{document}
\maketitle
\begin{abstract}
    \textbf{Objectives:} The rapid advancement of Multimodal Large Language Models (MLLMs) has significantly enhanced their reasoning capabilities, 
    enabling a wide range of intelligent applications. However, these advancements also raise critical concerns regarding privacy and ethics. 
    MLLMs are now capable of inferring the geographic location of images---such as those shared on social media or captured from street views---based solely on visual content, 
    thereby posing serious risks of privacy invasion, including doxxing, surveillance, and other security threats. 
    \textbf{Methods:} This study provides a comprehensive analysis of existing geolocation techniques based on MLLMs. 
    It systematically reviews relevant litera-ture and evaluates the performance of state-of-the-art visual reasoning models on geolocation tasks, 
    particularly in identifying the origins of street view imagery. 
    \textbf{Results:} Empirical evaluation reveals that the most advanced visual large models can successfully localize the origin of street-level imagery with up to 49\% 
    accuracy within a 1-kilometer radius. This performance underscores the models' powerful capacity to extract and utilize fine-grained geographic cues from visual data. 
    \textbf{Conclusions:} Building on these findings, the study identifies key visual elements that contribute to suc-cessful geolocation, such as text, 
    architectural styles, and environmental features. Furthermore, it discusses the potential privacy implications associated with MLLM-enabled geolocation and 
    discuss several technical and policy-based coun-termeasures to mitigate associated risks.
    Our code and dataset are available at \url{https://github.com/zxyl1003/MLLM-Geolocation-Evaluation}.
\end{abstract}    
\section{Introduction}
\label{sec:intro}

Multimodal large language models (MLLMs), as an emerging research hotspot in the field of artificial intelligence, 
use powerful large language models (LLMs) as their core processing units to perform multimodal tasks. 
MLLMs have demonstrated powerful capabilities in various fields such as image recognition and mathematical reasoning~\cite{Yin2024}. 
Due to their excellent performance in integrating multimodal information, as well as their excellent performance in tasks 
such as visual question answering and text-to-image world alignment, MLLMs have significant potential in geolocation tasks 
to provide more credible location predictions than traditional methods, which are essential for real-world applications 
such as remote sensing, disaster response, rescue, military, and autonomous navigation~\cite{firmansyah2024improving,ye2025icrossviewgeolocalizationnatural}. 
However, the power of MLLMs brings great convenience but also introduces new and potential privacy risks.

On one hand, MLLMs are capable of inferring the geographic location of an image solely based on its visual content, 
even in the absence of explicit geotags. This ability allows MLLMs to overcome the data dependency limitations faced by 
traditional image geolocation systems. On the other hand, such ``inference-based localization'' also implies that 
images uploaded by users—despite containing no geolocation metadata or textual information---may still be accurately geolocated 
by these models without the user's knowledge, thereby posing significant risks to personal privacy and the exposure of 
sensitive locations.

Moreover, most mainstream MLLMs are general-purpose models that lack dedicated constraints or ethical oversight mechanisms 
for handling geographic information. As a result, the risk of these models being misused for unauthorized image localization 
increases substantially. This risk is particularly pronounced in contexts such as social media, journalism, 
and military reconnaissance, where background elements in images may be automatically parsed by MLLMs and correlated 
with external knowledge bases, enabling the identification and tracking of specific individuals, facilities, 
or event locations. If exploited maliciously, such capabilities could pose serious threats to individual privacy, 
national security, and public order.

Therefore, a comprehensive investigation into the geolocation capabilities and potential risks associated with MLLMs is 
not only of academic interest but also of practical significance for the development of technical standards, 
privacy protection strategies, and ethical governance frameworks. 
Against this backdrop, this paper conducts a systematic evaluation of the state-of-the-art MLLMs in performing 
geolocation tasks using street-view imagery, and further discusses the challenges and risks that may arise in real-world 
applications.


\section{Related Work}
\label{sec:related_work}
Multimodal models can be broadly categorized into contrastive learning-based non-generative models and generative models~\cite{lin2025surveymechanisticinterpretabilitymultimodal}. 
Contrastive learning-based multimodal models, exemplified by CLIP~\cite{radford2021learning}, typically consist of a visual encoder and a text encoder, 
aligning images and text representations through contrastive learning. In contrast, generative multimodal models employ a multilayer perceptron (MLP) 
with several linear layers as a bridge to connect the visual encoder and the language model, enabling more integrated multimodal reasoning. 
At present, generative MLLMs have gained significant popularity, with prominent examples including the Qwen-VL model series~\cite{bai2025qwen25vltechnicalreport}, 
the DeepSeek-VL model~\cite{wu2024deepseekvl2mixtureofexpertsvisionlanguagemodels}, Gemini~\cite{deepmind2025gemini}, and OpenAI's family of multimodal models~\cite{openai2023gpt4vision}.

\subsection{Contrastive Learning-Based Multimodal Geolocation}
\label{sec:contrastive_learning_based_multimodal_geolocation}
StreetCLIP~\cite{haas2023learninggeneralizedzeroshotlearners} and GeoCLIP~\cite{cepeda2023geoclip} were among the first to introduce CLIP into the field of image Geolocation, 
both employing CLIP as the backbone to enable global-scale location inference. 
The key distinction lies in their alignment strategies: StreetCLIP aligns street-view images with textual 
descriptions containing city-level information, whereas GeoCLIP directly aligns images with geographic coordinates
(latitude and longitude). Building upon these ideas, the AddressCLIP model proposed by the Alibaba team~\cite{xu2024addressclip} 
associates images with human-readable address texts, though its study is limited to the San Francisco area. 
These con-trastive learning-based geolocation approaches primarily focus on adapting CLIP's zero-shot capabilities
to the geolocation domain through image-text pair pretraining. However, they fundamentally rely on correlations between 
multimodal data rather than a deep understanding of the visual content itself. As such, they lack comprehensive world 
knowledge and the capacity for spatial reasoning in a human-like manner—elements that are essential for robust and 
generalizable geolocation.

\subsection{Vision-Language Model-Based Multimodal Geolocation}
\label{sec:vision_language_model_based_multimodal_geolocation}
With the advancement of vision-language reasoning models, they have demonstrated strong capabilities in understanding, 
reasoning, and commonsense knowledge across a wide range of tasks, and are increasingly able to act as autonomous agents 
through tool use. Wang et al.~\cite{wang2024llmgeobenchmarkinglargelanguage} evaluated a series of multimodal models in image geolocation tasks, 
partially showcasing the geolocation inference capabilities of MLLMs. Their work also verified that fine-tuning or 
applying few-shot learning to MLLMs can effectively improve localization accu-racy, though the evaluation was limited 
to the national level rather than precise geographic coordinates. Jay et al.~\cite{jay2025evaluatingprecisegeolocationinference} not only assessed single-image localization 
capabilities on popular MLLMs, but also enabled access to external tools such as Street View and Google Lens. 
However, their results indicated that enabling tools can sometimes introduce additional noise into the reasoning process 
of MLLMs and does not always lead to better inference outcomes. Liu et al.~\cite{liu2024imagebasedgeolocationusinglarge} evaluated the impact of different prompt 
designs on MLLM geolocation results and explored the privacy challenges associated with MLLMs' localization capabilities. 
In addition, other related studies have evaluated and discussed the geolocation capabilities and ethical implications of 
MLLMs from various perspectives~\cite{huang2025vlmsgeoguessrmastersexceptional,mendes2024granularprivacycontrolgeolocation}.

A number of studies have also attempted to train MLLMs specifically for geolocation tasks. 
Li et al.~\cite{li2024georeasoner} proposed the GeoReasoner method, which employs a two-stage LoRA fine-tuning strategy to enhance the 
performance of MLLMs in geolocation. Notably, they introduced the novel concept of locatability to filter high-quality 
street-view images for constructing the training dataset. However, the training data lacked detailed reasoning chains; 
in the first stage, prompts were used to guide the MLLM to generate localization rationales, but the generated explanations 
were brief and did not reach the level of full reasoning chains. Anonymous~\cite{anonymous2025navig} proposed the NAVIG method, 
which constructed a high-quality but small-scale NAVICLUES dataset by mining commentary from expert GeoGuessr players 
on social media. This dataset was used to fine-tune MLLMs, and external tools such as OSM search and text retrieval were 
incorporated. The outputs of these tools were fed back into the model to improve geolocation accuracy. 
Yi et al.~\cite{yi2025geolocsftefficientvisualgeolocation} introduced GeoLocSFT, demonstrating that supervised fine-tuning (SFT) with only a few thousand high-quality 
``geo-captions'' can enable smaller MLLMs to achieve performance highly competitive with larger proprietary models. 
Song et al.~\cite{song2025geolocationrealhumangameplay} constructed a large-scale, human-annotated dataset, GeoComp, derived from real-world GeoGuessr gameplay, 
and proposed GeoCoT, a multi-step reasoning framework that mimics human thought processes to improve MLLM geolocation performance.

The aforementioned studies aim to evaluate or train MLLMs for street-view image geolocation. Street-view images typically contain 
a rich variety of visual elements. With the release of OpenAI's more powerful visual reasoning model, o3, 
researchers have sought to investigate whether MLLMs possess the ability to infer the locations of highly privacy-sensitive 
photos. Luo et al.~\cite{luo2025doxinglensrevealinglocationrelated} collected 50 highly private personal selfie images from the internet to evaluate o3's 
geolocation capabilities. The results demonstrated that o3 could easily infer geographic information from these privacy-sensitive 
images, with an accuracy significantly exceeding that of non-experts, thereby revealing serious risks of widespread privacy 
leakage. Furthermore, the study proposed GEOMINER, a collaborative attack framework designed to simulate adversaries enhancing 
MLLM geolocation inference by supplying contextual clues, thereby improving location prediction accuracy. 
The paper also explored possible defense mechanisms to mitigate the risk of such privacy breaches.
\section{Experiment Setting}
\label{sec:experiment_setting}
\subsection{Dataset}
\begin{figure*}[t]
    \centering
     \includegraphics[width=0.8\textwidth]{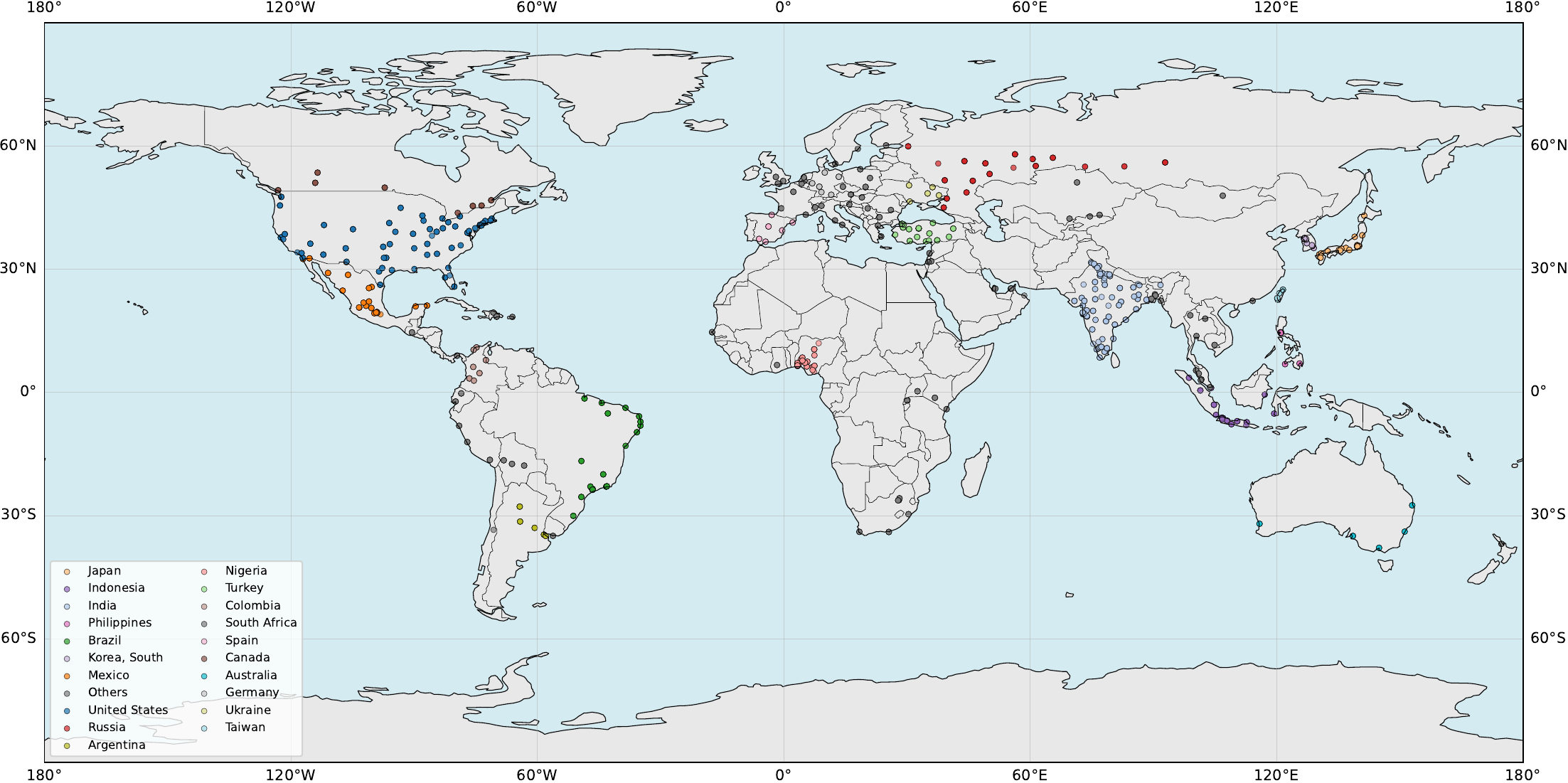}
     \caption{Global Distribution of Datasets.}
     \label{fig:1}
\end{figure*}

\begin{figure*}[t]
    \centering
     \includegraphics[width=0.8\textwidth]{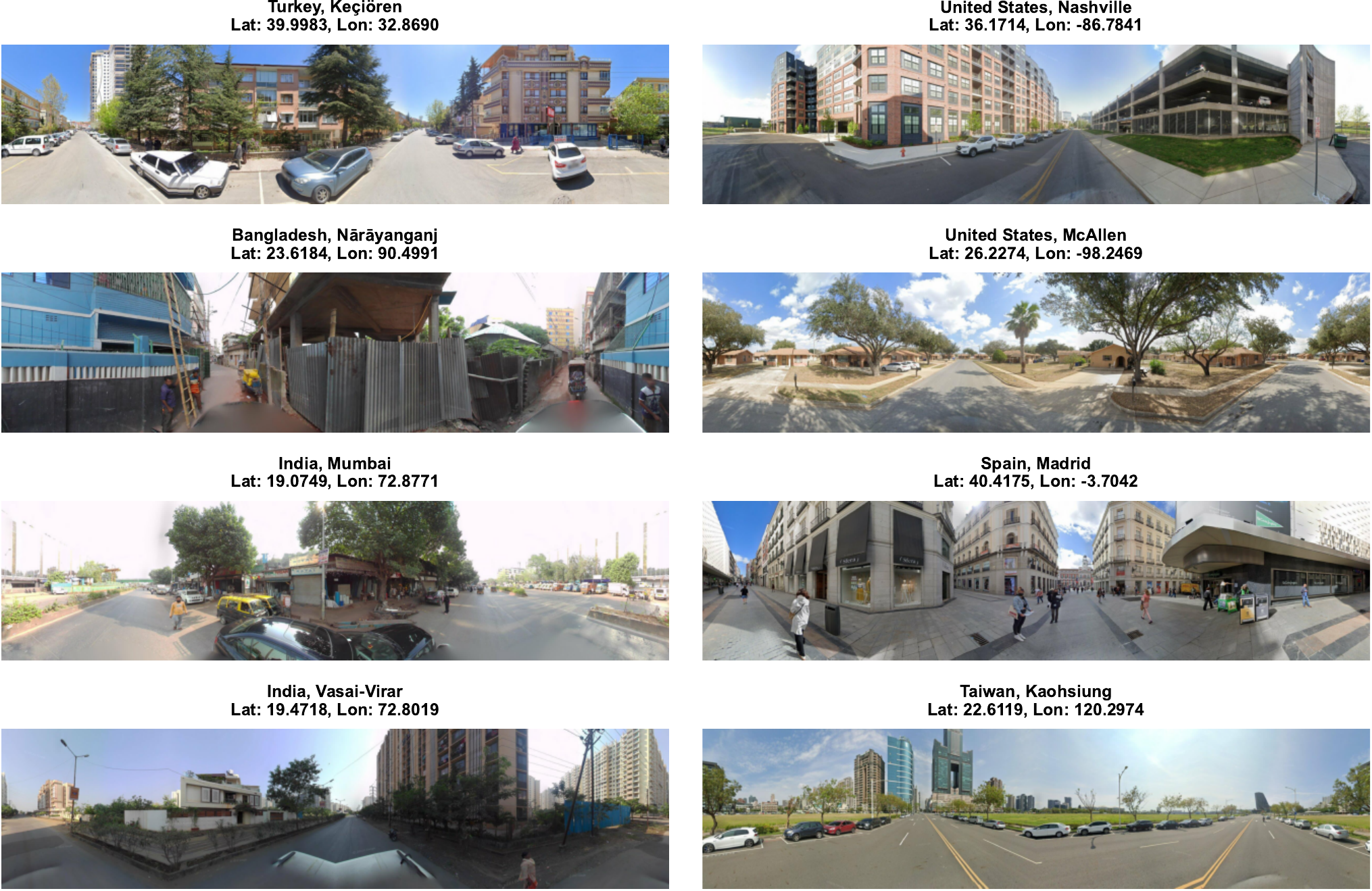}
     \caption{Sample Images of the Dataset.}
     \label{fig:2}
\end{figure*}

We obtained a global city dataset from Simplemaps.com, which contains 4,800 cities along with their corresponding latitude and 
longitude coordinates and population information. From this dataset, we selected the top 1,000 cities ranked by population. 
For each city, five sampling points were randomly generated within a 10 km radius of the city coordinates. 
Using the Google Street View API\footnote{https://developers.google.com/maps/documentation/streetview/}, street-view images were captured at four directions--$0^{\circ}$, $90^{\circ}$, $180^{\circ}$, and $270^{\circ}$--and 
stitched together to form panoramic images. Subsequently, the Google watermark located at the bottom of each image 
was cropped out. This process resulted in an evaluation dataset comprising 1,683 panoramic images. 
Figures~\ref{fig:1} and \ref{fig:2} illustrate the global distribution and sample images of the dataset, respectively. 
During the evaluation, due to API cost limitations, a subset of 300 panoramic images was randomly sampled from the full set of 1,683 images for model assessment.

\subsection{Model Evaluation}
In this study, we compare several state-of-the-art MLLMs with visual reasoning capabilities, 
including OpenAI's o1, Alibaba's Qwen-VL-Max and QvQ-Max, Anthropic's Claude-3.7-Sonnet-Thinking, 
ByteDance's Doubao-1.5-Thinking-Vision-Pro, Google's Gemini-2.5-Pro and Gemini-2.5-Flash, 
and Amazon's Nova-Pro-V1. Among them, the o1 model was evaluated using 38 images from the subset, 
while the other models were evaluated on the full 300-image subset. 
We designed a system prompt to guide the models to analyze visual clues step by step and to output 
their reasoning process. Each model ultimately produces a response in JSON format, containing the 
predicted latitude and longitude, as well as the inferred city and country.

We use two metrics to evaluate the model predictions---distance error and geographic score. 
The distance error is calculated using the Haversine formula:
\begin{equation}
    d = R\arctan^2\left(\sqrt{\mathrm{hav}\theta},\sqrt{1-\mathrm{hav}\theta}\right) \label{eq:1}
\end{equation}
where $R$ represents the average radius of the Earth, and $\mathrm{hav}\theta$ is computed as follows:
\begin{equation}
    \mathrm{hav}\theta = \sin^2\left(\frac{\Delta \phi}{2}\right) + \cos(\phi_1) \cos(\phi_2) \sin^2\left(\frac{\Delta \lambda}{2}\right) \label{eq:2}
\end{equation}
where $\Delta \phi$ and $\Delta \lambda$ represent the differences in latitude and longitude between the predicted and ground truth locations, respectively.

The GeoScore is derived from the game GeoGuessr\footnote{https://www.geoguessr.com/} and is calculated based on the distance error:
\begin{equation}
    \mathrm{geoscore} = 5000\times \exp\left(-\frac{d}{1492.7}\right)\label{eq:3}
\end{equation}
where $d$ refers to the distance error computed in the previous step, measured in kilometers. 
The GeoScore ranges from 0 to 5000, with higher values indicating greater proximity to the correct location.
  
\section{Experiment Results and Analysis}
\label{sec:experiment_results_and_analysis}
\subsection{Evaluation Results of MLLMs' Geolocation Capabilities}
The performance of different models is shown in Table~\ref{tab:1}. 
Although the o1 model demonstrated strong reasoning capabilities, 
it was evaluated on only 38 images; therefore, we include its results for reference but exclude it from further 
analysis. Among all models, we consider Gemini-2.5-Pro to have achieved the best overall performance 
across multiple metrics. Notably, it accurately localized images within a 1 km radius in 49\% of the cases. 
This indicates that current state-of-the-art MLLMs have, to a large extent, acquired global-scale geolocation 
capabilities. Such capabilities pose a serious threat to personal location privacy: an ordinary photo shared 
on social media, if containing sufficient visual information, may risk revealing geographic location. 
Unknowingly, the photos user publicly share may disclose our whereabouts, daily routines, 
or even personal identity. In more severe cases, these technologies could be exploited by malicious actors, 
resulting in threats such as doxxing or targeted surveillance.

\begin{table*}[t]
    \centering
    \caption{Testing results of different models on the evaluation dataset (300 images).}
    \begin{tabular}{@{}lcccccc@{}}
      \toprule
      & \multicolumn{2}{c}{geoscore} & \multicolumn{2}{c}{distance error} & \multicolumn{2}{c}{localization accuracy} \\
      \cmidrule(r){2-3} \cmidrule(lr){4-5} \cmidrule(l){6-7}
      & mean & std & mean & std & street ($\le 1$ km) & city ($\le 25$ km) \\
      \midrule
      o1 (38 images) & 4969.4 & 4998.4 & 10.1 & 0.5 & 68.4 & 97.4 \\
      Qwen-vl-max & 4398.2 & 4988.1 & 338.5 & 3.5 & 26.3 & 64.3 \\
      QvQ-max & 4455.8 & 4987.5 & 286.3 & 3.7 & 28 & 66.7 \\
      claude-3.7-sonnet-thinking & 4198.5 & 4979.5 & 523.6 & 6.1 & 26.3 & 58.3 \\
      doubao-1.5-thinking-vision-pro & 4594.9 & 4992.5 & 212.7 & 2.2 & 31.7 & 74.7 \\
      gemini-2.5-pro & \textbf{4725.2} & \textbf{4996.4} & \textbf{141.1} & \textbf{1.1} & \textbf{49.0} & \textbf{81.7} \\
      gemini-2.5-flash & 4623.2 & 4996.3 & 296.1 & 1.1 & 48.7 & 77.0 \\
      nova-pro-v1 & 3917.7 & 4658.8 & 859.2 & 105.5 & 20.7 & 49.7 \\
      \bottomrule
    \end{tabular}
    \label{tab:1}
\end{table*}

\begin{figure*}[t]
    \centering
    \includegraphics[width=0.8\textwidth]{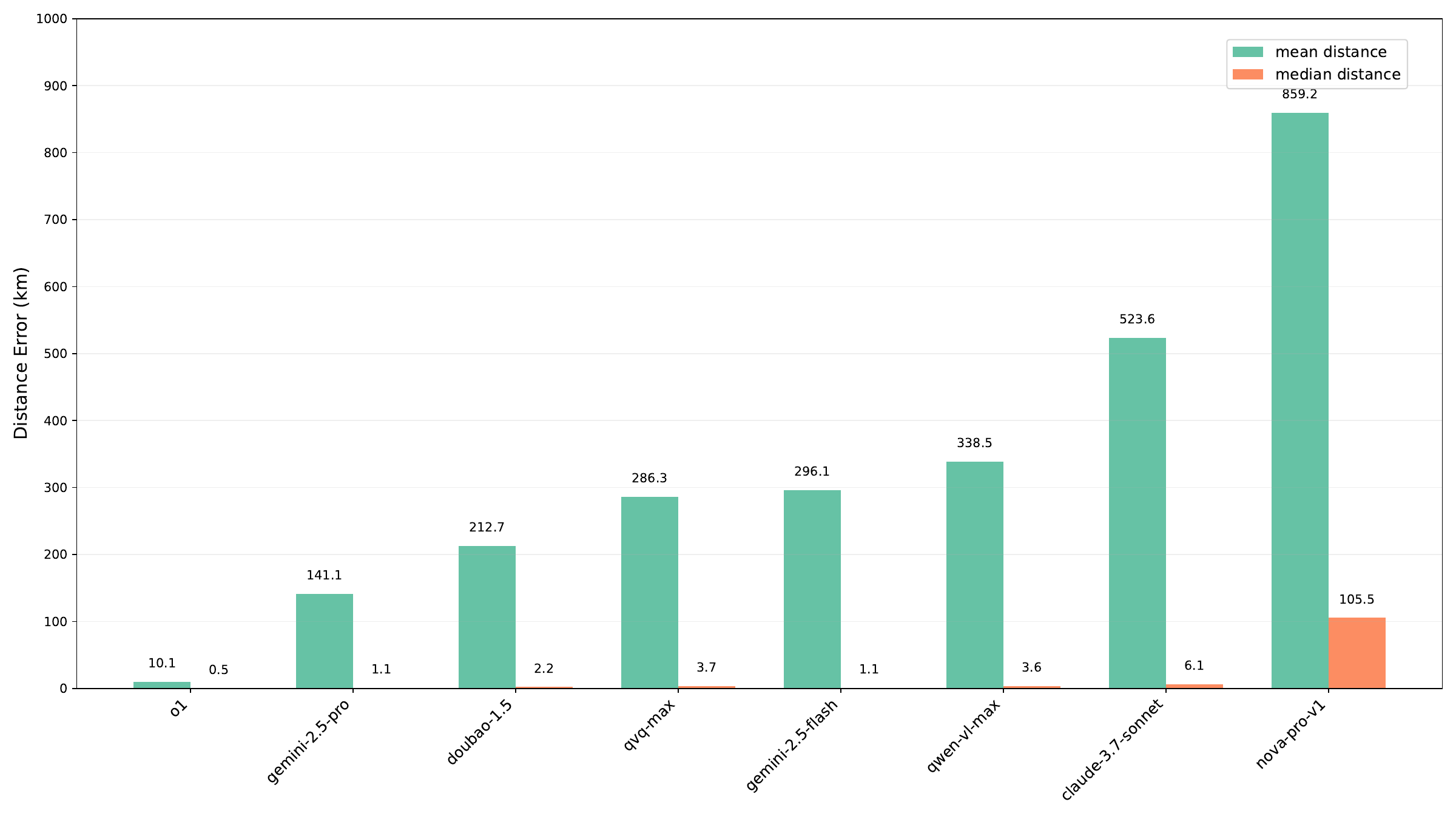}
    \caption{Visualization of Validation Results for Different Models.}
    \label{fig:3}
\end{figure*}

Figure~\ref{fig:3} presents a visualization of the mean and median localization errors for different models. 
It is evident that the mean values are significantly higher than the medians across all models. 
This suggests a bi-modal distribution in the localization results---models tend to perform either extremely well or extremely poorly, 
with relatively few predictions falling in the middle-accuracy range. This is an interesting phenomenon. 
We hypothesize that MLLMs, when performing geolocation, do not engage in a continuous, progressive reasoning process akin to human logical inference. 
Instead, their inference appears to rely on identifying ``decisive evidence'' within the image. 
That is, the model tends to locate one or more high-value geographic ``strong signals''; once such a signal is correctly recognized and matched, the localization result can be highly accurate. 
Conversely, when such strong signals are absent, the model must rely on more generic and ambiguous features to make a ``best guess,'' which is prone to large errors and significant geographic deviation. 
Moreover, geographic cues themselves are often inherently non-smooth. For example, 
if an MLLM recognizes a building as having ``Spanish colonial architecture,'' 
it may infer that the image was taken in Mexico City, Lima, or Madrid---but is unlikely to guess a random location 
200 kilometers northeast of Mexico City. In other words, geographic cues tend to produce either highly 
accurate predictions or highly erroneous ones, with few moderate outcomes.

This phenomenon reflects the discontinuous nature of MLLM reasoning. 
Unlike humans, whose spatial cognition tends to be continuous and stepwise, 
MLLMs operate based on discrete, high-dimensional feature matching, leading to a reasoning process that can 
appear abrupt and non-linear.

\subsection{Analysis of Key Visual Cues in Image GeoLocalization}
\begin{figure*}[t]
    \centering
    \includegraphics[width=0.8\textwidth]{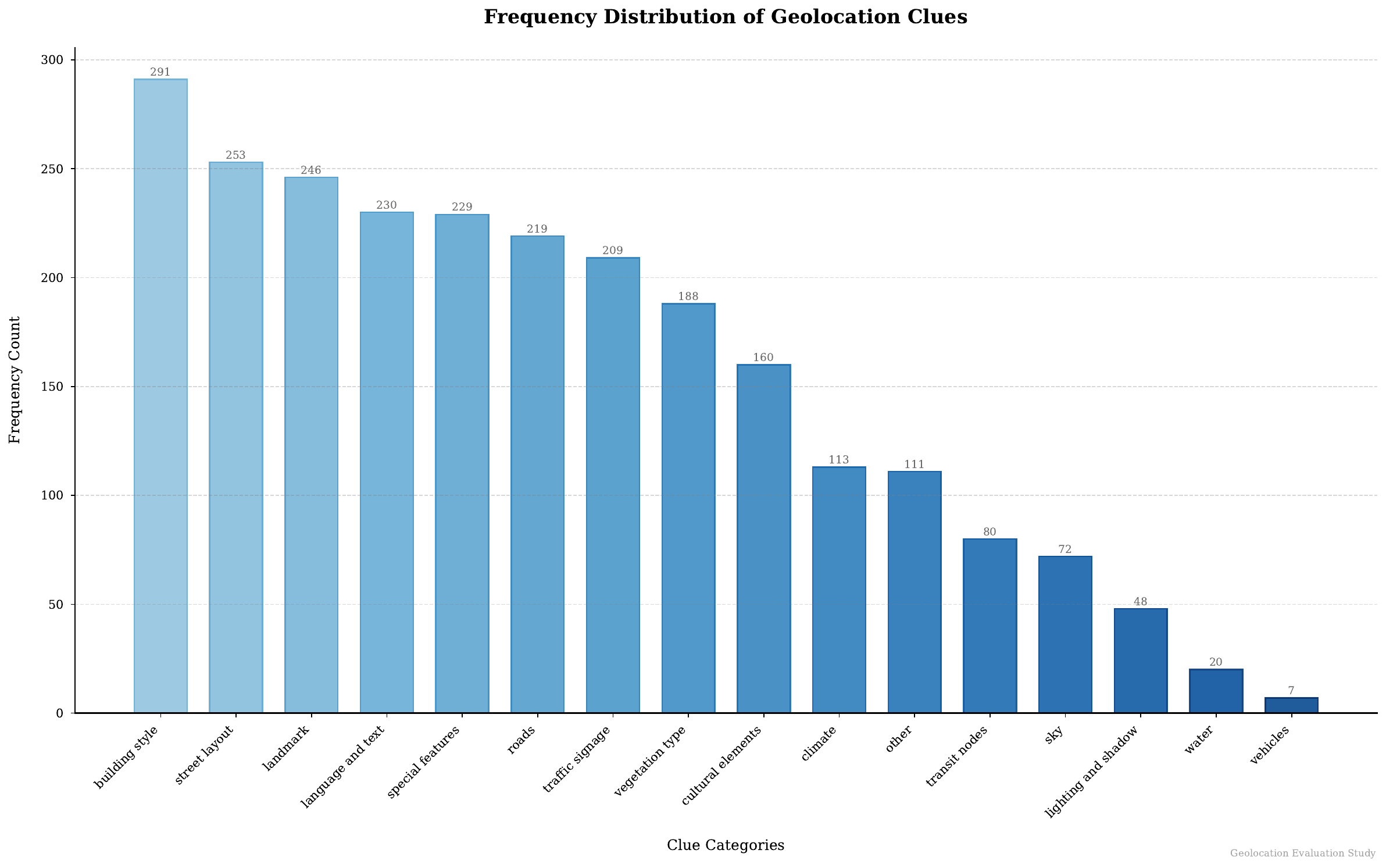}
    \caption{The Frequency of Different Cues Appearing During the Inference of the Google Gemini 2.5 Pro Model.}
    \label{fig:4}
\end{figure*}

\begin{figure*}[t]
    \centering
    \includegraphics[width=0.8\textwidth]{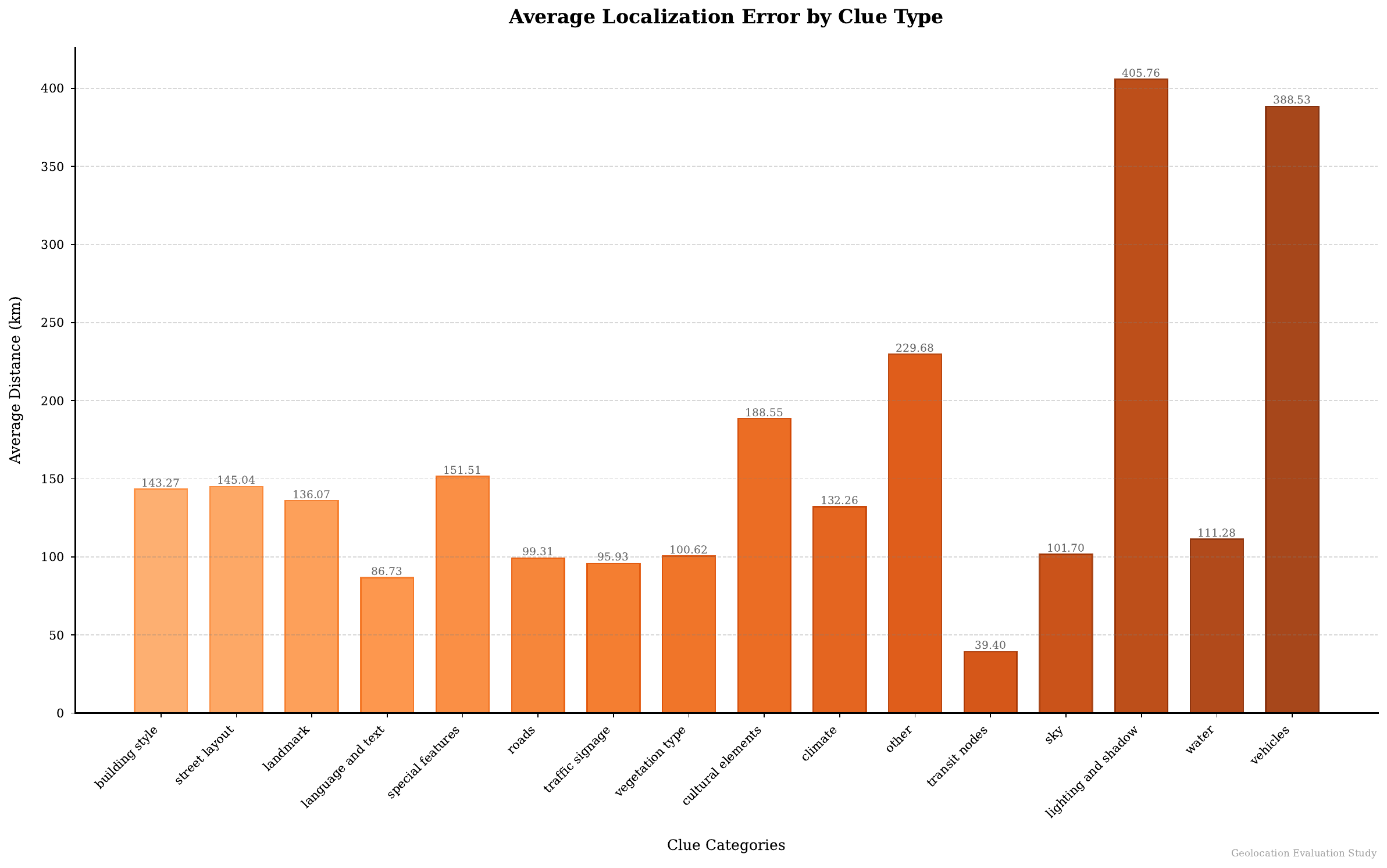}
    \caption{Visualization of Validation Results for Different Models.}
    \label{fig:5}
\end{figure*}

To identify which factors constitute “decisive evidence” for determining image locations, we analyzed the visual 
cues used by Google Gemini 2.5 Pro during its reasoning process. Specifically, we plotted the frequency of 
different types of visual elements and their corresponding average localization error. 
We first constructed a taxonomy of 16 types of visual elements, including: Building style, Street layout, 
Landmark, Language and Text, Special features, Roads, Traffic signage, Vegetation type, Cultural elements, 
Climate, Transit nodes, Sky, Lighting and Shadow, Water, Vehicles, and Other. 
A dedicated prompt was then designed to extract the model's reasoning chain from Gemini 2.5 Pro. 
This reasoning chain was subsequently fed into another LLM to classify the key visual elements mentioned 
during the inference process. The results are shown in Figures~\ref{fig:4} and \ref{fig:5}.

Figure~\ref{fig:4} illustrates the frequency with which each visual cue appears across the dataset. 
Among all types, Building style was the most frequently identified feature, appearing in 97\% of cases. 
In the remaining 3\%, the images were likely taken in enclosed or obstructed environments where building features were not visible. 
The least frequently used cue was related to Vehicles, likely due to two factors: vehicle features are often not uniquely region-specific, and license plates are blurred in Google Street View images. 
As shown in Figure~\ref{fig:5}, cues related to vehicles also resulted in large localization errors.

Figure~\ref{fig:5} shows the average distance error associated with each visual cue. 
Transit nodes (e.g., train stations, bus stops, airports) yielded the most accurate localization, 
as such infrastructure is typically geographically fixed and unique. 
The second most effective cue was Language and Text, which commonly appears on traffic signs, storefronts, and billboards, providing strong location-specific indicators. 
We define cues with an average error below 100 km as ``decisive evidence.''

This analysis reveals which types of information MLLMs most frequently rely on during geolocation reasoning, 
and which visual cues are most likely to expose the location of a photograph. 
By identifying these cues, users can better understand which elements in an image are most likely to compromise privacy. 
This enables more informed sharing practices---such as avoiding the upload of certain images to public platforms, 
or masking sensitive features prior to publication---and also provides valuable insights for the development of image privacy protection technologies.

\subsection{General Process of MLLM Image GeoLocation Reasoning}
\begin{figure*}[t]
    \centering
    \includegraphics[width=\textwidth]{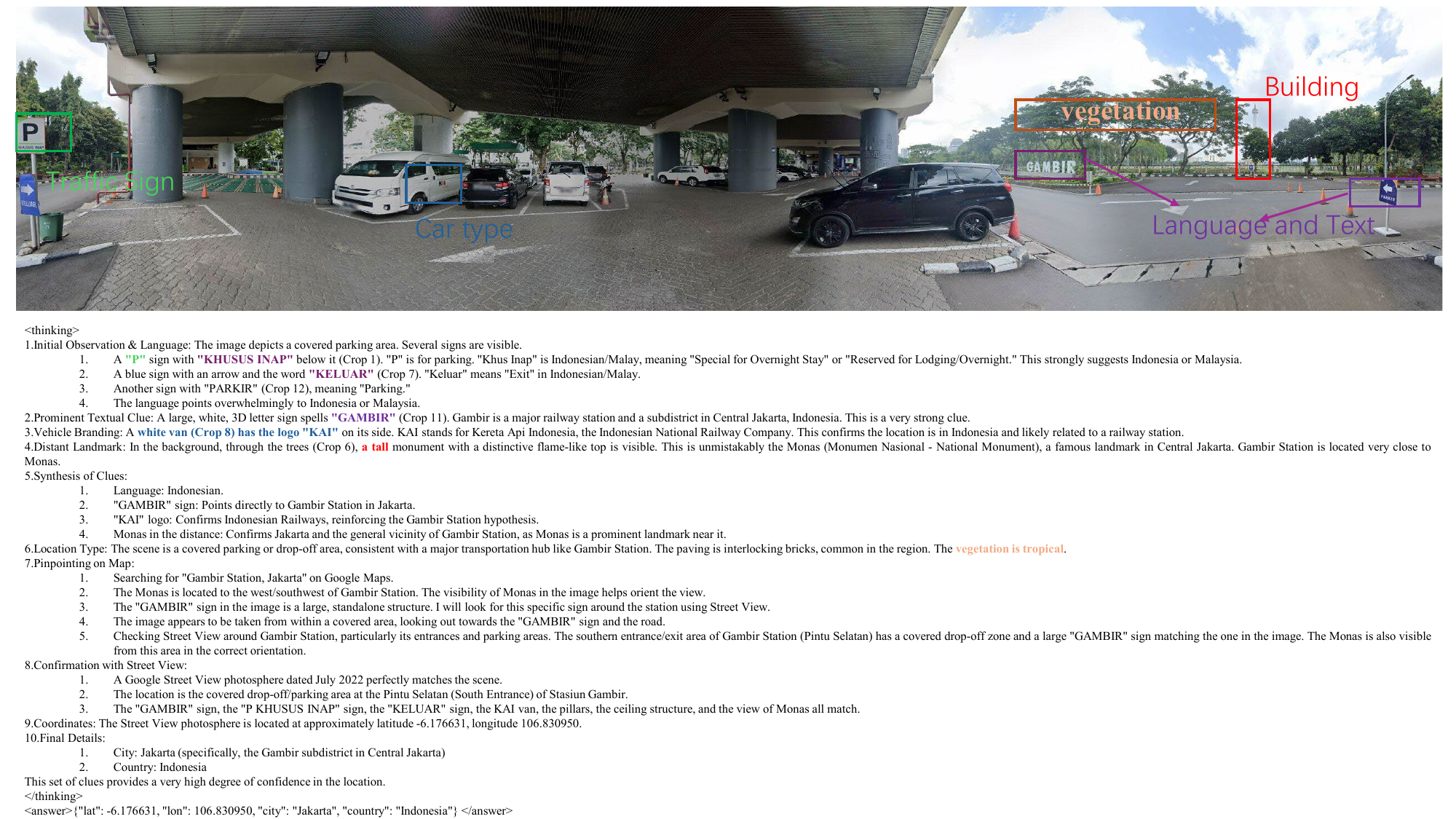}
    \caption{General Process of MLLM Image GeoLocation Reasoning.}
    \label{fig:6}
\end{figure*}

Figure~\ref{fig:6} illustrates the general process by which Gemini 2.5 Pro infers the location of an image, which can be divided into the following stages:

\begin{enumerate}
    \item Perception Stage: The model first identifies various visual elements within the image, including textual information (such as traffic signs, place names, and languages), 
    architectural structures, vehicle identifiers, vegetation types, terrain features, and more. 
    This stage combines image recognition and optical character recognition (OCR) capabilities.
    \item Analysis Stage: The extracted information is then correlated with the model's internal world knowledge. 
    For example, recognizing textual clues like ``KHUSUS INAP'' and ``KELUAR'' leads to the inference that the image may have been taken in Indonesia or Malaysia; 
    the place name ``GAMBIR,'' the railway company mark ``KAI,'' and the distant ``Monas National Monument'' suggest that the image was captured near Gambir Station in Jakarta, Indonesia. 
    Environmental details such as tropical plants and road materials are also analyzed to assist in localization.
    \item Spatial Verification and Fine Localization Stage: After forming an initial hypothesis, 
    the model in-vokes external geographic information tools, including the Google Maps and 
    Google Street View APIs, to perform comparative verification of the target area. 
    By matching structural features, signage, landmark ori-entations, and other details from specific 
    Street View perspectives, the model refines and confirms the im-age's location, 
    ultimately outputting precise latitude and longitude coordinates.
\end{enumerate}

From this process, it can be seen that the reasoning ability of MLLMs in geolocation tasks primarily derives from two aspects: 
first, the rich world knowledge acquired through large-scale multimodal pretraining, which enables the model to associate local visual cues with geographic information; 
and second, the tool-augmented mechanism, whereby external API tools are called upon as needed to supplement reasoning or validation. 
This capability highlights both the high practical value and the privacy-sensitive risks of current multimodal large language models (MLLMs) in visual spatial reasoning and real-world geolocation applications.

\subsection{Privacy Risks of MLLM Image GeoLocation}
\subsubsection{Potential Privacy Concerns}
The experiments described above demonstrate the powerful ability of MLLMs to perform image geolocation, 
which enables the unintended exposure of precise geographic locations through background elements, even when users do not explicitly provide location information. 
While this capability offers conveniences across various domains such as social media, law enforcement, military operations, and public opinion monitoring, 
it also raises serious privacy risks. These include tracking individuals’ activity trajectories, inferring the residences of public figures, 
locating and surveilling protest participants, and even facilitating malicious acts like doxxing. 
More alarmingly, the automation and scalability enabled by AI substantially lower the technical barriers, 
making large-scale, covert, and highly accurate geographic inference feasible. Users thus face unprecedented risks of ``being located'' without their awareness or ability to defend themselves.

\subsubsection{Privacy Protection Strategies}
To mitigate location privacy leakage caused by MLLMs, interventions should be pursued on technical, 
regulatory, and user-control fronts. Technically, image obfuscation methods---such as blurring or masking 
distinctive visual cues like text and traffic signs---can be applied before image sharing to disrupt model localization capabilities. 
On the regulatory side, stronger legal frameworks and industry standards are essential, exemplified by explicit restrictions on geographic information as sensitive data under GDPR and PIPL, 
as well as requirements for AI system explainability and user informed consent. 
From the user perspective, individuals should have the right to know and control how their content is analyzed; 
platforms should offer privacy modes or location obfuscation tools. 
Overall, a multi-pronged approach with platform accountability at the forefront is the key to addressing AI-enabled geolocation threats and safeguarding location privacy.
\section{Conclusion and Future Works}
\label{sec:conclusion_and_future_works}
\subsection{Conclusion}
This study constructed a geolocation evaluation dataset consisting of 1,683 panoramic images. 
Based on this dataset, we conducted a systematic assessment of the performance of state-of-the-art MLLMs on geolocation tasks. 
The results demonstrate that, leveraging both their intrinsic world knowledge and external tool integration, 
current leading geolocation models possess remarkably strong geolocation capabilities, achieving up to a 49\% success rate in accurately localizing street-view images within a 1 km radius. 
In addition, we observed a ``bimodal'' phenomenon in MLLMs' geolocation performance and provided an explanation for its underlying mechanism, offering new insights for the further development of MLLMs' spatial perception abilities. 
We also identified which visual cues most effectively guide MLLMs in correctly inferring the precise geographic coordinates of images. 
Furthermore, this study discusses the privacy risks posed by MLLM geolocation capabilities. 
In an era of global communication and widespread social media, visual information about individuals and organizations is ubiquitous. 
Current MLLMs can infer sensitive location information from just a few images. Malicious actors may exploit this capability for doxxing through online media, raising serious security and privacy concerns.

\subsection{Future Works}
This study provides a systematic evaluation of the performance of state-of-the-art MLLMs on image geolo-cation tasks and discusses the associated privacy risks. However, several limitations remain. 
First, the street-view dataset used for evaluation is relatively small in scale. Although the collected data is globally distributed, only 300 images were used for evaluation due to cost constraints. 
Second, the most advanced OpenAI visual reasoning model was evaluated on only 38 images, which may lead to an unfair assessment due to insufficient data. 
Finally, this work does not systematically evaluate the impact of tool usage on MLLM performance. Since models like Gemini 2.5 Pro and o3 inherently support tool integration, 
whereas others such as Qwen do not, we did not construct external tool environments for models lacking this capability.

In future work, we aim to develop a specialized geolocation model with performance comparable to state-of-the-art MLLMs at a lower cost by leveraging open-source models such as Qwen-2.5-VL-32B. 
We plan to adopt a two-stage training paradigm similar to DeepSeek-R1: the first stage involves cold-start fine-tuning of the MLLM using high-quality chain-of-thought data to enhance the model's 
spatial cognition for image localization; the second stage uses large-scale image-coordinate pair data for reinforcement learning to improve localization accuracy. 
To achieve this, a large-scale geolocation dataset will be constructed by collecting street-view data globally via the Google Street View API, 
with corresponding chain-of-thought datasets gen-erated using Gemini 2.5 Pro. To mitigate the observed bimodal (``double-peak'') phenomenon, 
we intend to build a hierarchical chain-of-thought reasoning framework progressing from macro to micro levels. 
This stepwise structured reasoning is expected to reduce cascading errors caused by single erroneous clues. 
To further improve MLLM performance, we will emulate Gemini 2.5 Pro to construct an agent dataset aimed at enhancing the model's tool invocation capability. 
Additionally, compared to single street-view images, videos contain richer visual cues and spatial constraints. 
Building on image localization, we will explore MLLM-based geolocation of videos in future research.
{
    \small
    \bibliographystyle{ieeenat_fullname}
    \bibliography{main}
}


\end{document}